\documentclass[10pt,twocolumn,letterpaper]{article}

\usepackage[pagenumbers]{cvpr}
\usepackage{colortbl}

\usepackage{cuted} 

\usepackage{currfile} 

\usepackage{caption} 
\definecolor{cvprblue}{rgb}{0.21,0.49,0.74}
\usepackage[pagebackref,breaklinks,colorlinks,allcolors=cvprblue]{hyperref}
\usepackage{xcolor}

\def\confName{CVPR}
\def\confYear{2027}

\title{SReWAM: JEPA-Written TTT-Memory for One-Step World-Action Models}
\author{
\mbox{
Hang Yu \enspace
Wenrui Bao \enspace
Kunyang Li \enspace
Anh Dao \enspace
Yuzhang Shang$^{*}$ \enspace
Mubarak Shah$^{*}$
}\\[0.2em]
Institute of Artificial Intelligence, University of Central Florida\\[0.2em]
}

\begin{document}
\maketitle

\begin{abstract}
World-action models (WAMs) leverage predictive visual representations for robotic manipulation, yet iterative action generation incurs substantial inference overhead. Existing one-step methods primarily modify action generation rather than enriching historical conditioning, which becomes particularly important without iterative refinement. We propose \textbf{SReWAM}, a sparse representation-driven memory framework that provides persistent historical conditioning for one-step action generation. SReWAM constructs memory keys and values directly from pretrained V-JEPA representations and accumulates historical information in fast weights through self-supervised test-time training (TTT), without explicitly storing past observations. An observation-aligned write-then-read mechanism enables policy-conditioned memory retrieval within a single denoising evaluation. Sparse integration into selected action-expert layers limits additional overhead while keeping pretrained backbones frozen in simulation. On LIBERO-Plus, SReWAM improves one-step success rates from 77.1\% to 78.9\% on JEPA-WAM and from 85.3\% to 86.0\% on a JEPA-adapted $\pi_{0.5}$ backbone. Sparse integration achieves comparable performance to dense memory with 75\% fewer additional trainable parameters. Shared-checkpoint ablations demonstrate benefits from online updates and cross-observation persistence. On a real-robot can-to-plate task, SReWAM achieves 30/30 successful trials in the original setting and 10/10 under a held-out tabletop appearance shift.
\end{abstract}

\begin{figure*}[t]
    \centering
    \includegraphics[width=\textwidth,trim=0.2cm 6.4cm 3.7cm 0.3cm,clip]{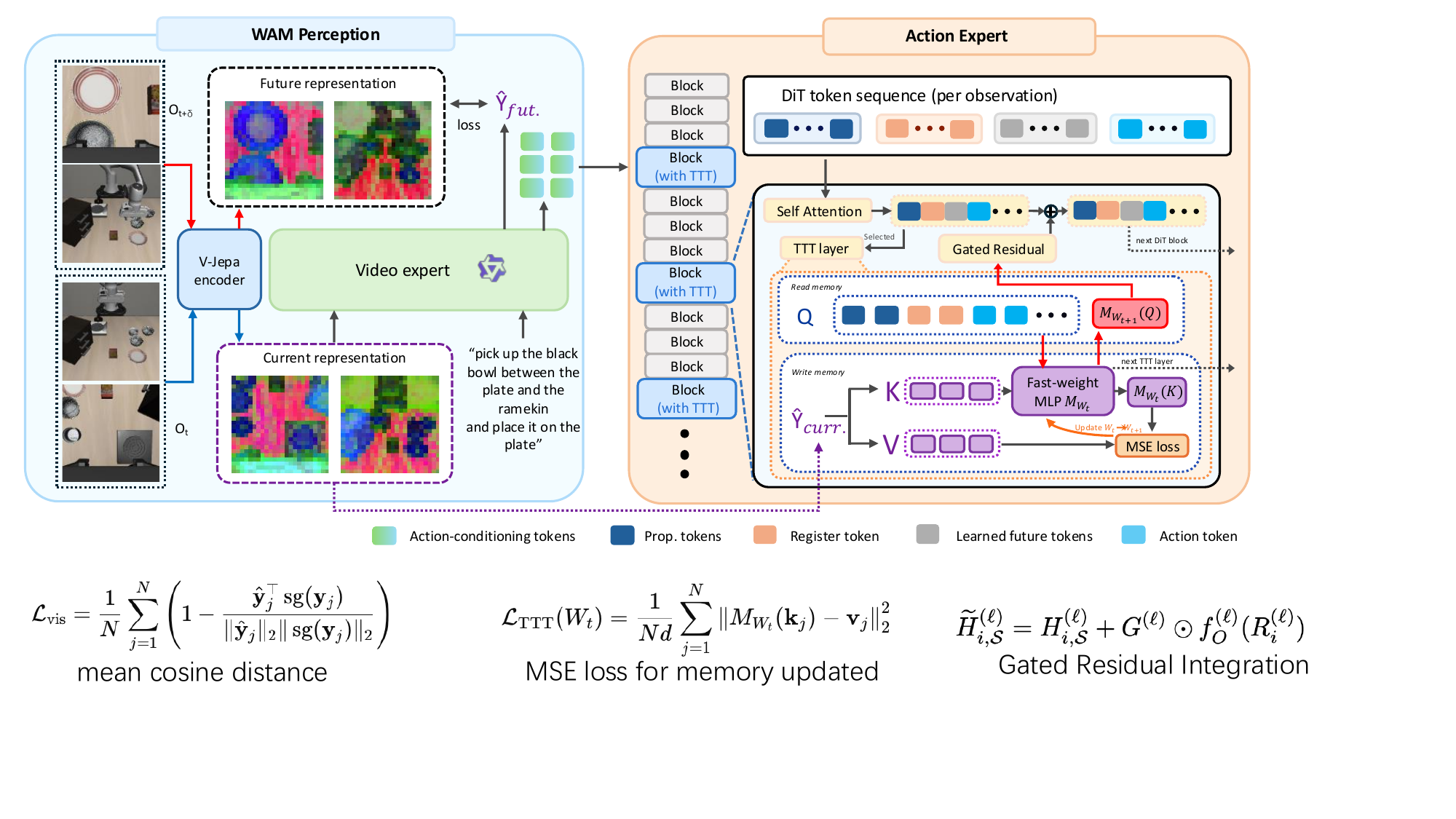}    
\caption{Overview of SReWAM. Action-conditioning tokens provide multimodal context to the action expert through cross-attention, while current JEPA visual representations drive online fast-weight memory updates. Selected post-attention policy tokens retrieve the updated memory through gated residual connections. Sparse TTT wrappers provide additional historical conditioning for one-step action generation.}
    \label{fig:architecture}
\end{figure*}

\section{Introduction}
\label{sec:introduction}

World-action models (WAMs) couple predictive world modeling with action generation, enabling robotic policies to anticipate future dynamics rather than rely solely on current observations~\cite{cen2025worldvla,bi2026motus,lin2026jepa}. Recent WAMs streamline world modeling through latent-space predictions instead of reconstructing future images~\cite{yuan2026fast,huang2026foresight}, while JEPA-based WAMs condition action experts on predictive visual representations~\cite{lin2026jepa,assran2025v,mur2026v}. Yet action generation remains iterative: diffusion or flow-matching action experts often require multiple denoising evaluations per observation, incurring substantial inference overhead.

Reducing action generation to a single denoising step eliminates iterative refinement. Existing one-step methods primarily change how actions are generated, using distillation, velocity calibration, or structured action priors~\cite{prasad2024consistency,wang2024one,jia2024score,luan2026snapflow,akbari2026flash,wang2026anystep,su20261step,zhang2026oswameffectiveworldaction}. These methods focus on the generation process rather than explicitly enriching the action expert's historical conditioning. This distinction matters under a one-step budget: multi-step diffusion and flow-matching policies progressively refine action estimates across evaluations~\cite{chi2024diffusionpolicyvisuomotorpolicy,ze20243d,liu2025rdt,black2024pi_0,lin2026jepa}, whereas one-step policies generate an action chunk in a single evaluation. We therefore explore whether richer conditioning can improve one-step action prediction rather than further approximating multi-step generation.

Interaction history is a promising source of richer conditioning. Robotic manipulation is sequential, and current observations can miss past interactions, especially when objects are occluded or earlier states are no longer visible. Memory-augmented policies preserve history through stored observations or compressed visual tokens~\cite{shi2026memoryvla,torne2026mem,yang2026memorywam}. However, repeatedly processing historical observations adds inference overhead, potentially reducing the efficiency gains of one-step generation. Test-time training (TTT) instead encodes sequential information into dynamically updated fast weights, preserving memory across observations without storing or reprocessing the full history~\cite{sun2024learning,liu2026t,wang2026test,feng2026wam}. This raises a question: \emph{can predictive JEPA representations already available in WAMs support efficient persistent fast-weight memory for one-step action generation?}

To address this question, we propose \textbf{SReWAM}, a sparse representation-driven memory framework for one-step world-action models. Rather than relying on action-expert hidden states for memory writing or explicitly storing historical visual tokens, SReWAM directly constructs memory keys and values from pretrained V-JEPA representations of the current observation. These pretrained visual representations provide a memory-writing signal aligned with the predictive feature space used by the world model. Through self-supervised TTT updates, the fast-weight memory accumulates information across observations, while post-attention policy tokens retrieve the updated memory through gated residual connections. This explicit JEPA-driven writing and policy-conditioned retrieval design provides complementary historical conditioning for one-step action generation without maintaining an explicit observation history.

However, online fast-weight updates and memory retrieval introduce additional computational overhead, potentially offsetting the efficiency benefits of one-step inference. To address this trade-off, SReWAM sparsely integrates TTT modules into selected action-expert layers rather than throughout the entire network. The pretrained perception, world-modeling, and action-generation backbone remains frozen, while memory-related parameters are optimized using the flow-matching objective. At inference time, each observation triggers one fast-weight update per selected memory module, followed by immediate retrieval during action generation. This observation-aligned design provides persistent historical conditioning without additional distillation or velocity calibration.

We evaluate SReWAM on LIBERO-Plus and RoboTwin 2.0 across different JEPA-based WAM backbones. On LIBERO-Plus, SReWAM improves one-step success rates from 77.1\% to 78.9\% on JEPA-WAM and from 85.3\% to 86.0\% on the JEPA-adapted $\pi_{0.5}$ backbone. Sparse memory integration achieves comparable performance to dense integration with 75\% fewer trainable parameters. Further ablations examine memory-writing strategies, online updates, persistence, and inference efficiency. Our main contributions are summarized as follows:

\begin{itemize}
    \item We propose \textbf{explicit JEPA representation-driven memory writing}, which directly constructs memory keys and values from pretrained V-JEPA visual representations for self-supervised fast-weight updates, enabling persistent historical conditioning without explicitly storing past observations.

    \item We develop a \textbf{sparse, observation-aligned TTT integration} strategy that enables persistent memory updates and retrieval within one-step action generation, while preserving the pretrained backbone and limiting additional computational overhead.

    \item We demonstrate the \textbf{performance--efficiency trade-off} of the proposed design across JEPA-based WAM backbones, showing improved one-step manipulation performance and comparable accuracy to dense memory integration with substantially fewer trainable parameters.
\end{itemize}

\begin{figure*}[t]
    \centering
    \includegraphics[
        width=\textwidth,
        trim=0.1cm 2cm 0.1cm 11.5cm,
        clip
    ]{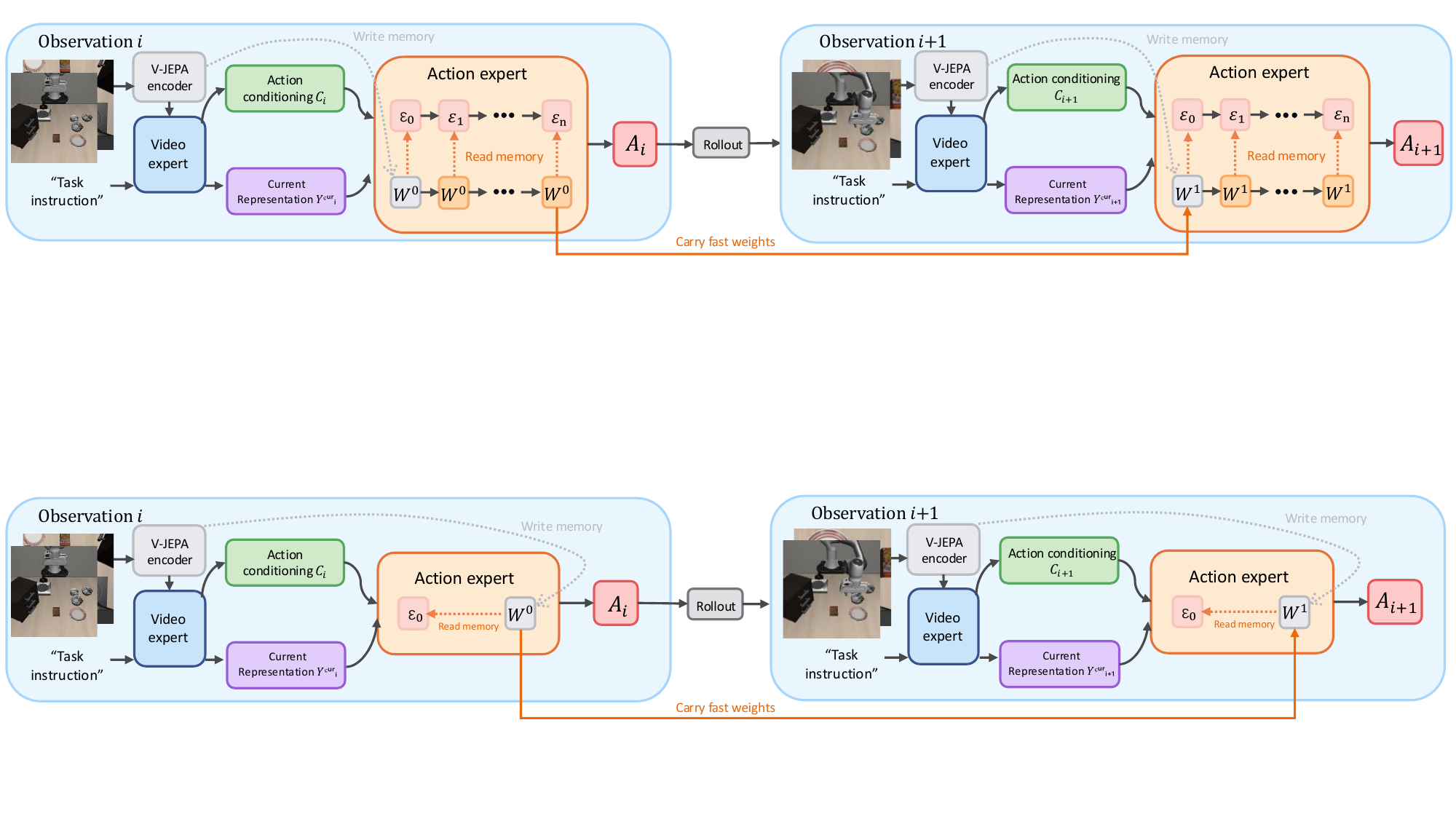}
    \caption{Observation-aligned memory updates during one-step action generation. At observation $i$, current JEPA representations update the persistent fast-weight memory, which is immediately queried for action prediction. The updated memory is carried to observation $i+1$ and further adapted using the new visual information. Here, $\epsilon_0$ denotes the initial action noise, while $W^0$ and $W^1$ schematically represent the effective memory states after the respective observation-level updates, rather than the fixed slow parameters.}
    \label{fig:inference}
\end{figure*}

\section{Related Work}
\label{sec:related_work}

\subsection{Predictive World-Action Models}

World-action models (WAMs) integrate predictive world modeling with robot action generation~\cite{cen2025worldvla,bi2026motus,zhou2026tau_0,guo2026unified}. Existing approaches leverage video prediction or latent world modeling to capture environment dynamics and provide future conditioning for manipulation~\cite{hu2024video,guo2026video,kim2026cosmos,yuan2026fast,zhao2026faster,huang2026foresight,chen2026lawam}. Joint-embedding predictive architectures (JEPAs) further enable prediction directly in representation space~\cite{assran2025v,mur2026v}. Recent JEPA-based robot policies exploit predictive visual representations for world modeling and action conditioning~\cite{lin2026jepa,zhang2026vjepapolicybuildingeffective,sun2026vla}. In contrast, SReWAM directly reuses current V-JEPA representations as memory-writing signals, enabling persistent historical conditioning across observations.

\subsection{Efficient One-Step Action Generation}

Diffusion- and flow-matching-based robot policies typically generate actions through iterative denoising~\cite{chi2024diffusionpolicyvisuomotorpolicy,ze20243d,liu2025rdt,black2024pi_0,intelligence2025pi_}. Recent methods reduce inference cost through consistency or diffusion distillation, adaptive denoising, velocity calibration, and structured action priors~\cite{prasad2024consistency,wang2024one,jia2024score,luan2026snapflow,akbari2026flash,wang2026anystep,su20261step,zhang2026oswameffectiveworldaction}. These approaches primarily modify the action-generation process rather than explicitly maintaining historical information as conditioning. SReWAM instead explores a complementary direction by enriching one-step action generation with persistent historical conditioning, without additional distillation or velocity calibration.

\subsection{Memory-Augmented Robot Policies and Test-Time Training}

Memory-augmented robot policies preserve historical information through temporal conditioning, perceptual memory, compressed visual tokens, and event-driven or memory-guided mechanisms~\cite{koo2026hamlet,shi2026memoryvla,torne2026mem,yang2026memorywam,yang2026eventvla}. Test-time training (TTT) offers an alternative by encoding sequential information into dynamically updated fast weights~\cite{sun2024learning,behrouz2026titans,zhang2026test}. Recent robotic approaches explore observation-grounded fast-weight memory~\cite{liu2026t}, long-context adaptation through TTT~\cite{jiang2026robottt}, and residual historical memory~\cite{wang2026test}. Related work also investigates video-driven memory adaptation in WAMs~\cite{feng2026wam} and persistent adaptation of latent world dynamics~\cite{zhang2026jepa}.

Unlike approaches that construct memory through learned vision-language interfaces~\cite{liu2026t}, SReWAM directly forms memory keys and values from pretrained V-JEPA representations. Current visual features drive self-supervised fast-weight updates, while post-attention policy tokens retrieve the updated memory through gated residual connections. By combining observation-aligned write-then-read updates with sparse integration into the action expert, SReWAM provides persistent historical conditioning under a single denoising step while limiting additional computational overhead.

\section{Method}
\label{sec:method}

\subsection{Framework Overview}
\label{sec:framework_overview}

As illustrated in Fig.~\ref{fig:architecture}, we introduce \emph{SReWAM}, which integrates sparse test-time training (TTT) memory into JEPA-based WAMs for one-step action generation. The pretrained vision-language backbone produces action-conditioning tokens $C_i$ from action-placeholder hidden states, providing visual, language, and task context as cross-attention keys and values. The action expert processes proprioceptive, learned register, future latent, and noisy action tokens, which serve as cross-attention queries. Proprioceptive tokens encode robot states, while action tokens incorporate noisy actions and flow time. Future latent tokens are learned placeholders, and only action-token outputs are decoded into predicted flow velocities.

SReWAM employs an explicit JEPA representation-driven memory mechanism that separates visual memory writing from policy-conditioned retrieval. Specifically, current V-JEPA visual representations provide keys and values for self-supervised fast-weight updates, while post-attention state, register, and action tokens query the updated memory through gated residual connections. Unlike the standard cross-attention pathway, memory keys and values are constructed directly from V-JEPA representations rather than action-conditioning tokens $C_i$. Future latent tokens receive no direct memory residuals. The fast weights persist across observations and reset at episode boundaries, enabling historical conditioning without explicitly storing past visual tokens. We focus on the Wrapper-Explicit (WE) configuration, with other variants detailed in the supplementary material.

\subsection{Representation-Driven Fast-Weight Memory}
\label{sec:representation_memory}

\paragraph{Memory Construction.}
SReWAM explicitly leverages current visual representations from the pretrained V-JEPA encoder as memory-writing signals, rather than relying on action-expert hidden states. These representations lie in the predictive feature space used by the world model, allowing memory updates to reuse pretrained visual information without introducing an additional visual memory encoder.

At observation $i$, let $Y_i^{\mathrm{cur}}$ denote the current V-JEPA visual features. For a TTT module at action-expert block $\ell$, memory keys and values are constructed as:
\begin{equation}
\begin{aligned}
    \bar{Y}_i &= \operatorname{L2Norm}(Y_i^{\mathrm{cur}}),\\
    K_i^{(\ell)} &= \operatorname{RoPE}_{s_i^{(\ell)}}
        \bigl(\operatorname{RMSNorm}(\bar{Y}_i)\bigr),\\
    V_i^{(\ell)} &= \bar{Y}_i,
\end{aligned}
\end{equation}

where $s_i^{(\ell)}$ is a per-sample temporal counter maintained independently by each TTT module. It starts at zero for each episode and increments once per valid fast-weight update. All key and query tokens within an update share the same RoPE position. The keys incorporate temporal positional information, while the values retain the normalized visual representations. Memory writing uses current JEPA features rather than predicted future representations.

Memory queries are constructed from post-attention policy hidden states:
\begin{equation}
    Q_i^{(\ell)}
    = \operatorname{RoPE}_{s_i^{(\ell)}}
      \bigl(f_Q^{(\ell)}(H_{i,\mathcal{S}}^{(\ell)})\bigr),
\end{equation}
where $\mathcal{S}$ selects the proprioceptive state, register, and action tokens, and $f_Q^{(\ell)}$ denotes a learned query projection. This asymmetric construction separates JEPA representation-driven memory writing from policy-conditioned retrieval.

\paragraph{Online Fast-Weight Update.}
Each TTT module maintains an independent two-layer GELU MLP with parameters $F_i^{(\ell)}=W_\ell+\Delta_i^{(\ell)}$, where $W_\ell$ denotes learned slow weights and $\Delta_i^{(\ell)}$ represents the persistent fast-weight offset.

Given the previous memory state, we define a self-supervised regression objective:
\begin{equation}
    \mathcal{L}_{\mathrm{mem},i}^{(\ell)}
    = \frac{1}{N_K d_m}
    \left\|
        M_{F_{i-1}^{(\ell)}}(K_i^{(\ell)})
        - V_i^{(\ell)}
    \right\|_F^2,
\end{equation}
where $N_K$ is the number of memory key tokens and $d_m$ is the memory feature dimension. The fast-weight offset is updated through a gated gradient step:
\begin{equation}
\begin{aligned}
    \Delta_i^{(\ell)}
    ={}& \Delta_{i-1}^{(\ell)}
    - \operatorname{softplus}(\alpha_\ell)
      \,g_i^{(\ell)} \\
    &\cdot \nabla_{F_{i-1}^{(\ell)}}
       \mathcal{L}_{\mathrm{mem},i}^{(\ell)},
\end{aligned}
\label{eq:prewam_inner_update}
\end{equation}
where $\alpha_\ell$ is a learnable learning-rate parameter and $g_i^{(\ell)}\in(0,1)$ is a scalar update gate computed from pooled, normalized policy features using a two-layer SiLU network with sigmoid output. The updated memory state is $F_i^{(\ell)}=W_\ell+\Delta_i^{(\ell)}$, allowing visual information to accumulate across observations through persistent fast weights.

\paragraph{Memory Retrieval and Integration.}
Unlike read-before-write memory schemes, SReWAM first updates the fast weights using the current JEPA representations before performing policy-conditioned retrieval. The post-attention policy queries then retrieve information from the updated memory:
\begin{equation}
    R_i^{(\ell)}
    = M_{F_i^{(\ell)}}(Q_i^{(\ell)}).
\end{equation}
The retrieved features are integrated into the policy hidden states through a channel-wise gated residual:
\begin{equation}
    \widetilde{H}_{i,\mathcal{S}}^{(\ell)}
    = H_{i,\mathcal{S}}^{(\ell)}
      + G^{(\ell)}
      \odot f_O^{(\ell)}(R_i^{(\ell)}),
\end{equation}
where $f_O^{(\ell)}$ is the output projection and $G^{(\ell)}=\tanh(a_\ell)$ is a learned channel-wise residual scale initialized near zero.

The residual is applied between attention and feed-forward operations only to the selected policy tokens, leaving future latent tokens without direct memory residuals. This observation-aligned write-then-read mechanism enables the action expert to immediately access memory updated with the current observation, alongside information accumulated from previous interactions, within a single evaluation.

\subsection{Sparse Memory Integration for One-Step Action Generation}
\label{sec:persistent_action}

\paragraph{Sparse TTT Integration.}
To limit the computational and parameter overhead of persistent memory, SReWAM inserts TTT wrappers into selected action-expert blocks rather than throughout the entire network. Specifically, four wrappers are placed at blocks 4, 8, 12, and 16 (one-based indexing) of the 16-block JEPA-WAM action expert, with the same positions used in the 18-block JEPA-adapted $\pi_{0.5}$ action expert. These positions are predetermined rather than selected through a layer-placement search. This sparse placement distributes memory access across the action expert while avoiding the cost of equipping every block with an independent fast-weight module. Compared with dense integration, it substantially reduces additional trainable parameters while preserving the pretrained backbone computations.

\paragraph{Observation-Aligned Memory and Inference.}
As illustrated in Fig.~\ref{fig:inference}, fast-weight memory persists across observations and resets at episode boundaries. Under one-step inference, each observation triggers one self-supervised fast-weight update per selected TTT module, followed immediately by memory retrieval and action generation using a single flow-matching denoising step. This write-then-read mechanism allows the action expert to access both newly incorporated visual information and accumulated historical memory within the same denoising evaluation, without additional denoising steps.

For multi-step inference, memory is updated only during the first denoising evaluation of each observation and remains fixed throughout subsequent steps. The action expert repeatedly retrieves information from the updated memory while refining its action prediction. Thus, memory updates are aligned with incoming observations rather than the denoising budget. The same schedule is applied to both JEPA-WAM and the JEPA-adapted $\pi_{0.5}$ backbone, with further implementation details provided in the supplementary material.

\subsection{Training Objective and Optimization}
\label{sec:prewam_training}

Unless otherwise specified, SReWAM is trained on temporally ordered windows of $T=16$ observations sampled at a stride of eight, spanning a nominal 128-frame temporal context. Each observation $i$ is paired with a ground-truth action chunk $\mathbf{a}_i \in \mathbb{R}^{H \times d_a}$, where $H$ is the action horizon and $d_a$ is the action dimension.

For the JEPA-WAM backbone, we sample $\boldsymbol{\epsilon}_i\sim\mathcal{N}(\mathbf{0},\mathbf{I})$ and $u_i\sim\operatorname{Beta}(1.5,1)$, and define
\begin{equation}
\begin{aligned}
    \tau_i &= \frac{0.999-u_i}{0.999},\\
    \mathbf{x}_i^{\tau_i}
    &= (1-\tau_i)\boldsymbol{\epsilon}_i
       + \tau_i\mathbf{a}_i,\\
    \mathbf{v}_i^\star
    &= \mathbf{a}_i-\boldsymbol{\epsilon}_i.
\end{aligned}
\end{equation}
The JEPA-adapted $\pi_{0.5}$ backbone follows its original flow-matching time and velocity conventions, as detailed in the supplementary material.

Given observation-dependent conditioning $\mathbf{c}_i$ and the updated fast-weight memory, the action expert predicts the flow velocity $\hat{\mathbf{v}}_i$. Let $m_{ih}\in\{0,1\}$ indicate whether action step $h$ is valid. The outer training objective is the masked flow-matching loss:
\begin{equation}
    \mathcal{L}_{\mathrm{FM}}
    = \frac{
        \displaystyle
        \sum_{i=1}^{T}
        \sum_{h=1}^{H}
        \sum_{d=1}^{d_a}
        m_{ih}
        \left(
            \hat{v}_{ihd}
            - v_{ihd}^{\star}
        \right)^2
    }{
        \displaystyle
        d_a
        \sum_{i=1}^{T}
        \sum_{h=1}^{H}
        m_{ih}
    }.
\label{eq:prewam_flow_loss}
\end{equation}

The inner self-supervised memory objective updates the fast weights, while the outer flow-matching objective optimizes the TTT-related slow parameters and the introduced register embeddings through differentiable memory updates. In simulation experiments, the pretrained perception, world-modeling, and action-generation backbone remains frozen. We employ eight-observation truncated backpropagation through time (TBPTT), retaining fast-weight values while detaching their computation graphs between segments. Fast-weight offsets are initialized to zero at the beginning of each independent training window.

\begin{table*}[t]
\centering

\caption{Main results on LIBERO-Plus. All metrics report success rates (\%, higher is better). SReWAM improves one-step success rates by 1.8 and 0.7 percentage points on JEPA-WAM and JEPA-adapted $\pi_{0.5}$, respectively, demonstrating the benefit of memory augmentation under a single denoising step. Results are grouped by robot-policy pretraining. JEPA-based baselines use publicly released checkpoints evaluated under our pipeline. Best results within each group are bolded, and our primary one-step configuration is shaded in gray.
}
\label{tab:libero_plus}
\small
\setlength{\tabcolsep}{3.0pt}
\renewcommand{\arraystretch}{0.95}
\begin{tabular}{@{}lcccccccc@{}}
\toprule
\textbf{Method}
& \textbf{Camera}
& \textbf{Robot}
& \textbf{Language}
& \textbf{Light}
& \textbf{Back.}
& \textbf{Noise}
& \textbf{Layout}
& \textbf{Avg.} $\uparrow$ \\
\midrule
\multicolumn{9}{@{}l}{\textbf{Part I: Without robot-policy pretraining}} \\
VLA-Adapter~\cite{wang2026vla}
& 36.2 & 37.9 & 74.6 & 70.6 & 76.1 & 58.0 & 69.7 & 60.4 \\
RoVLA~\cite{luo2026rovla}
& 58.4 & 36.3 & \textbf{92.9} & 95.6 & \textbf{95.0} & 80.9 & 73.0 & 76.0 \\
ResVLA~\cite{zhong2026noise}
& 49.8 & \textbf{59.9} & 88.5 & 90.5 & 94.9 & 76.8 & \textbf{79.0} & 77.1 \\
\multicolumn{9}{@{}l}{JEPA-WAM~\cite{lin2026jepa}} \\
\hspace{1.6em}-- 4-step denoising
& 77.7 & 57.2 & 66.4 & 92.9 & 92.4 & 83.5 & 74.6 & 77.8 \\
\hspace{1.6em}-- 1-step denoising
& 77.3 & 55.0 & 64.9 & 92.9 & 91.2 & 83.5 & 74.8 & 77.1 \\
\midrule
\multicolumn{9}{@{}l}{\textit{SReWAM (Ours): Wrapper-Explicit}} \\
\hspace{1.6em}-- 4-step denoising
& 78.7 & 58.8 & 65.8 & 94.7 & 93.7 & \textbf{84.9} & 75.7 & \textbf{78.9} \\
\rowcolor[gray]{0.94}
\hspace{1.6em}\textbf{-- 1-step denoising}
& \textbf{78.8} & 57.6 & 66.1 & \textbf{96.1} & 93.9 & 84.4 & 75.4 & \textbf{78.9} \\
\midrule
\multicolumn{9}{@{}l}{\textbf{Part II: With robot-policy pretraining}} \\
VLA-JEPA~\cite{sun2026vla}
& 63.3 & 67.1 & 85.4 & 95.6 & 93.6 & 66.3 & 85.1 & 79.5 \\
PokeVLA~\cite{zheng2026pokevla}
& \textbf{84.7} & 46.1 & 84.8 & 94.6 & 82.6 & 89.8 & 77.2 & 80.0 \\
ABot-M0~\cite{yang2026abot}
& 60.4 & 67.9 & 86.4 & 96.2 & 91.6 & 86.4 & 82.6 & 81.6 \\
Cosmos-Policy~\cite{kim2026cosmos}
& 75.8 & 63.3 & 81.7 & 96.5 & 88.9 & \textbf{92.7} & 82.2 & 83.0 \\
$\pi_{0.5}$~\cite{intelligence2025pi_}
& 69.4 & 75.3 & 82.6 & 96.7 & 96.8 & 84.3 & 86.2 & 84.5 \\
Being-H0.7~\cite{luo2026being}
& 82.0 & 59.0 & 82.8 & 97.8 & 90.0 & 93.5 & 88.5 & 84.8 \\
\multicolumn{9}{@{}l}{$\pi_{0.5}$ + JEPA Obj.~\cite{lin2026jepa}} \\
\hspace{1.6em}-- 4-step denoising
& 66.0 & \textbf{82.6} & 85.5 & 97.6 & \textbf{97.2} & 85.2 & 88.0 & 86.0 \\
\hspace{1.6em}-- 1-step denoising
& 64.5 & 80.4 & 86.8 & 97.4 & 95.4 & 84.4 & 88.3 & 85.3 \\
\midrule
\multicolumn{9}{@{}l}{\textit{$\pi_{0.5}$ + SReWAM (Ours): Wrapper-Explicit}} \\
\hspace{1.6em}-- 4-step denoising
& 66.9 & 82.3 & \textbf{87.1} & 97.0 & 96.7 & 85.2 & \textbf{88.9} & \textbf{86.3} \\
\rowcolor[gray]{0.94}
\hspace{1.6em}\textbf{-- 1-step denoising}
& 65.5 & \textbf{82.6} & 86.8 & \textbf{98.2} & \textbf{97.2} & 83.3 & 88.4 & 86.0 \\
\bottomrule
\end{tabular}
\end{table*}

\begin{table}[t]
\centering
\caption{Success rates (\%) on four selected RoboTwin 2.0 tasks under the Clean setting. Published baselines ($^\dagger$) are taken from JEPA-WAM~\cite{lin2026jepa}, with averages recomputed over the four tasks. The published $\pi_{0.5}$ baseline uses 10-step denoising, while locally evaluated JEPA-adapted models use the indicated denoising steps.}
\label{tab:robotwin}
\scriptsize
\setlength{\tabcolsep}{2.5pt}
\renewcommand{\arraystretch}{1.05}
\begin{tabular*}{\columnwidth}{@{}l@{\extracolsep{\fill}}ccccc@{}}
\toprule
\textbf{Method}
& \shortstack{\textbf{Adjust}\\\textbf{Bottle}}
& \shortstack{\textbf{Dump Bin}\\\textbf{Bigbin}}
& \shortstack{\textbf{Shake}\\\textbf{Bottle}}
& \shortstack{\textbf{Press}\\\textbf{Stapler}}
& \textbf{Avg.} $\uparrow$ \\
\midrule
RDT-1B$^\dagger$~\cite{liu2025rdt}
& 81 & 64 & 74 & 41 & 65.00 \\
$\pi_0^\dagger$~\cite{black2024pi_0}
& 90 & 83 & 97 & 62 & 83.00 \\
$\pi_{0.5}^\dagger$~\cite{intelligence2025pi_}
& 98 & 95 & 99 & 66 & 89.50 \\
\midrule
$\pi_{0.5}$ + JEPA Obj.~\cite{lin2026jepa}
& & & & & \\
\hspace{1em}-- 4-step
& 99 & 95 & 100 & 1 & 73.75 \\
\hspace{1em}-- 1-step
& 98 & 82 & 100 & 40 & 80.00 \\
\midrule
\multicolumn{6}{@{}l}{\textit{$\pi_{0.5}$ + SReWAM (WE)}} \\
\rowcolor[gray]{0.94}
\hspace{1em}\textbf{-- 1-step}
& 95 & 85 & 100 & 65 & \textbf{86.25} \\

\bottomrule
\end{tabular*}
\end{table}

\begin{table*}[t]
\centering
\caption{Ablation of TTT memory integration and writing strategies on LIBERO-Plus using the JEPA-WAM backbone without large-scale robot-policy pretraining. Inline (I) and Wrapper (W) denote dense and sparse integration, respectively; Explicit (E) and Implicit (I) denote current V-JEPA and policy-hidden-state memory writing. Best results are highlighted in bold, and the primary Wrapper-Explicit one-step configuration is shaded in gray.}
\label{tab:memory_design_ablation}
\small
\setlength{\tabcolsep}{3.0pt}
\renewcommand{\arraystretch}{0.95}
\begin{tabular}{@{}lcccccccc@{}}
\toprule
\textbf{Method}
& \textbf{Camera}
& \textbf{Robot}
& \textbf{Language}
& \textbf{Light}
& \textbf{Back.}
& \textbf{Noise}
& \textbf{Layout}
& \textbf{Avg.} $\uparrow$ \\
\midrule

\multicolumn{9}{@{}l}{\textbf{Inline-Implicit (II)}} \\
\hspace{1.6em}-- 4-step denoising
& \textbf{79.3} & 57.1 & 65.6 & 94.4 & 93.6 & 84.0 & 75.5 & 78.5 \\
\hspace{1.6em}-- 1-step denoising
& 78.3 & 57.0 & 65.7 & 94.5 & 93.0 & 84.4 & 75.1 & 78.3 \\

\midrule
\multicolumn{9}{@{}l}{\textbf{Inline-Explicit (IE)}} \\
\hspace{1.6em}-- 4-step denoising
& 78.0 & 57.6 & 66.4 & 95.1 & \textbf{94.0} & 84.3 & \textbf{76.1} & 78.8 \\
\hspace{1.6em}-- 1-step denoising
& 79.2 & 58.1 & \textbf{66.5} & 95.9 & 92.6 & 83.9 & 75.3 & 78.8 \\

\midrule
\multicolumn{9}{@{}l}{\textbf{Wrapper-Implicit (WI)}} \\
\hspace{1.6em}-- 4-step denoising
& 78.4 & 57.3 & 66.1 & 95.2 & 93.2 & 84.5 & 73.6 & 78.3 \\
\hspace{1.6em}-- 1-step denoising
& 78.5 & 57.3 & 65.9 & 94.6 & 92.5 & 84.1 & 73.6 & 78.1 \\

\midrule
\multicolumn{9}{@{}l}{\textbf{Wrapper-Explicit (WE)}} \\
\hspace{1.6em}-- 4-step denoising
& 78.7 & \textbf{58.8} & 65.8 & 94.7 & 93.7 & \textbf{84.9} & 75.7 & \textbf{78.9} \\
\rowcolor[gray]{0.94}
\hspace{1.6em}\textbf{-- 1-step denoising}
& 78.8 & 57.6 & 66.1 & \textbf{96.1} & 93.9 & 84.4 & 75.4 & \textbf{78.9} \\

\bottomrule
\end{tabular}
\end{table*}

\section{Experiments}
\label{sec:experiments}

\subsection{Experimental Setup}
\label{sec:experimental_setup}

\paragraph{Benchmarks.}
We evaluate SReWAM on LIBERO-Plus~\cite{fei2025libero} and RoboTwin 2.0~\cite{chen2025robotwin}. LIBERO-Plus covers seven perturbation categories, with 10,030 episodes per configuration; we report category-wise success and their unweighted mean. RoboTwin evaluation includes four selected Clean-setting bimanual tasks, with 100 episodes each. Full protocols are provided in the supplementary material.

\paragraph{Backbones.}
Part I uses JEPA-WAM~\cite{lin2026jepa}, with a V-JEPA 2.1 encoder~\cite{mur2026v} and a 16-block DiT action expert, without large-scale robot-policy pretraining. Part II uses a JEPA-adapted $\pi_{0.5}$ checkpoint~\cite{intelligence2025pi_,lin2026jepa} with an 18-block action expert. Both backbones remain frozen, with only TTT-related parameters and register embeddings trained.

\paragraph{Memory Variants and Inference.}
We compare sparse Wrapper and dense Inline integration, each using Explicit writing from V-JEPA representations or Implicit writing from post-attention DiT states. All variants retrieve memory through state, register, and action tokens. Part I evaluates all four variants, while Part II focuses on Wrapper-Explicit (WE). For JEPA-WAM, WE uses 76.66M trainable parameters versus 307M for Inline-Explicit (IE). We evaluate one- and four-step denoising, resetting memory at episode boundaries without additional distillation.

\subsection{Main Results on LIBERO-Plus}
\label{sec:libero_plus_results}

Table~\ref{tab:libero_plus} shows that SReWAM (WE) improves one-step success from 77.1\% to 78.9\% (+1.8 points) on JEPA-WAM, surpassing its four-step baseline (77.8\%). On the JEPA-adapted $\pi_{0.5}$ backbone, one-step success increases from 85.3\% to 86.0\% (+0.7 points). Paired bootstrap confidence intervals are reported in supplementary Table~\ref{tab:paired_bootstrap}.

Four-step improvements are smaller: 77.8\% to 78.9\% (+1.1 points) on JEPA-WAM and 86.0\% to 86.3\% (+0.3 points) on $\pi_{0.5}$. Both backbones thus show numerically larger gains under one-step inference. Notably, one-step SReWAM matches four-step success on JEPA-WAM and approaches it on $\pi_{0.5}$ (86.0\% vs. 86.3\%), suggesting that persistent memory can complement action generation under a limited denoising budget.

Improvements vary across perturbations. Under one-step inference, robot-initialization success increases by 2.6 and 2.2 points on JEPA-WAM and JEPA-adapted $\pi_{0.5}$, respectively, while background perturbations improve by 2.7 and 1.8 points. However, sensor-noise success decreases by 1.1 points on the stronger backbone, indicating setting-dependent rather than uniform robustness gains.

\subsection{Results on RoboTwin 2.0}
\label{sec:robotwin_results}

Table~\ref{tab:robotwin} reports results on four selected Clean-setting bimanual tasks. SReWAM (WE) uses 50 demonstrations per task (200 total), training only TTT-related parameters and register embeddings with a frozen backbone. Its one-step average reaches 86.25\%, versus 80.00\% for our evaluation of the released JEPA-adapted $\pi_{0.5}$ checkpoint. Press Stapler shows the largest gain (40\% to 65\%), while Adjust Bottle decreases from 98\% to 95\%. The published ten-step $\pi_{0.5}$ baseline achieves 89.50\%.

These results support frozen-backbone memory augmentation for multi-view bimanual manipulation. However, task-specific memory training prevents attributing the 6.25-point gain solely to persistence; shared-checkpoint ablations in Table~\ref{tab:memory_ablation} provide more direct evidence.

\subsection{Inference Efficiency}

Table~4 includes memory-update and retrieval latency. One-step SReWAM (WE) achieves 78.9\% success at 163.79 ms E2E latency, compared with 77.8\% at 191.81 ms for four-step JEPA-WAM, but remains slower than one-step JEPA-WAM (146.79 ms). Compared with dense IE, sparse WE achieves comparable success (78.9\% vs. 78.8\%) with 75\% fewer trainable parameters and lower latency (163.79 vs. 175.02 ms). One-step WE matches four-step success (78.9\%) with 33.7\% lower latency. Thus, sparse integration reduces, but does not eliminate, online memory overhead.

\subsection{Ablation Studies}
\label{sec:ablation_results}

\paragraph{Online Updates and Persistence.}
Table~\ref{tab:memory_ablation} isolates memory behavior using a shared WE checkpoint: Slow Only disables online updates; Reset performs observation-level updates but clears fast weights before the next observation; Full preserves them across observations. Success increases from 77.12\% (Slow Only) to 77.93\% (Reset) and 78.90\% (Full), yielding gains of 0.81 and 0.97 points, respectively. The separately trained Static Adapter achieves 77.20\%. Reset versus Full supports the contribution of cross-observation persistence beyond current-observation adaptation; confidence intervals appear in supplementary Table~\ref{tab:paired_bootstrap}.

\paragraph{Counterfactual Memory Sensitivity.}
Using the same WE checkpoint, we modify historical fast-weight offsets after four observation-level updates while holding current observations and policy inputs fixed. Real History achieves 78.28\% success, compared with 77.33\% for Cleared History and 76.48\% for Swapped History. Real History numerically outperforms both alternatives across all four LIBERO suites, with the largest Real-to-Swapped difference on Goal tasks (+4.81 points). Unlike full-episode persistence ablations, this intervention modifies memory at an intermediate observation. These exploratory results indicate sensitivity to historical memory content without establishing statistical significance. Full results appear in supplementary Table~\ref{tab:counterfactual_memory}.

\paragraph{Writing Source and Integration.}
Table~\ref{tab:memory_design_ablation} compares Explicit JEPA-based and Implicit DiT-based memory writing. Under one-step inference, Explicit writing improves success by 0.8 points with Wrapper integration (78.1\% to 78.9\%) and 0.5 points with Inline integration (78.3\% to 78.8\%). Four-step gains are 0.6 and 0.3 points, respectively. Dense IE provides no consistent advantage over sparse WE despite using four times as many TTT-equipped blocks. These results favor direct JEPA representation reuse and sparse integration, although representation semantics are not isolated from other architectural differences.

\paragraph{Training Context Length.}
Table~\ref{tab:context_length} shows that increasing training context from 16 to 128 frames (2--16 sampled observations) improves episode-level success from 76.72\% to 77.64\%. Goal tasks gain the most (69.74\% to 72.44\%), while Object tasks slightly decline (85.15\% to 84.91\%), suggesting task-dependent benefits from longer contexts. Unlike Table~\ref{tab:libero_plus}, Table~\ref{tab:context_length} reports overall episode-level success rather than the unweighted mean across perturbation categories.

\begin{table}[t]
\centering
\caption{Inference efficiency on LIBERO-Plus. One-step SReWAM (WE) achieves higher success with lower end-to-end latency than four-step JEPA-WAM. All latency measurements include online memory updates where applicable. Our primary one-step configuration is highlighted in gray.
}
\label{tab:inference_latency}
\small
\setlength{\tabcolsep}{3pt}
\renewcommand{\arraystretch}{1.05}
\begin{tabular*}{\columnwidth}{
@{}l@{\extracolsep{\fill}}cccc@{}
}
\toprule
\textbf{Method}
& \textbf{Steps}
& \textbf{Expert (ms)} $\downarrow$
& \textbf{E2E (ms)} $\downarrow$
& \textbf{Success} $\uparrow$ \\
\midrule
JEPA-WAM
& 4 & 62.31 & 191.81 & 77.8 \\
JEPA-WAM
& 1 & \textbf{16.43} & \textbf{146.79} & 77.1 \\
\midrule
SReWAM (WE)
& 4 & 116.65 & 247.18 & \textbf{78.9} \\
SReWAM (IE)
& 1 & 70.72 & 175.02 & 78.8 \\
\rowcolor[gray]{0.94}
\textbf{SReWAM (WE)}
& \textbf{1} & 30.86 & 163.79 & \textbf{78.9} \\
\bottomrule
\end{tabular*}
\vspace{3pt}
\parbox{\columnwidth}{
\footnotesize
\textbf{Speed--accuracy gain:}
\textbf{1.17$\times$ faster E2E inference} and \textbf{+1.1 pp success} compared with four-step JEPA-WAM.
}
\end{table}

\begin{table}[t]
\caption{Effect of training context length on LIBERO-Plus performance. Observations are sampled every eight frames. Total reports episode-level success across 10,030 evaluation episodes.}
\centering
\label{tab:context_length}
\small
\setlength{\tabcolsep}{4.5pt}
\renewcommand{\arraystretch}{0.95}
\begin{tabular}{@{}lccccc@{}}
\toprule
\textbf{Context}
& \textbf{Long} $\uparrow$
& \textbf{Goal} $\uparrow$
& \textbf{Object} $\uparrow$
& \textbf{Spatial} $\uparrow$
& \textbf{Total} $\uparrow$ \\
\midrule
16  & 68.44 & 69.74 & \textbf{85.15} & 84.10 & 76.72 \\
32  & 68.24 & 71.86 & 84.39 & 84.18 & 77.04 \\
64  & 68.72 & 71.48 & 85.07 & \textbf{84.64} & 77.35 \\
\rowcolor[gray]{0.94}
128 & \textbf{69.11} & \textbf{72.44} & 84.91 & 84.55 & \textbf{77.64} \\
\bottomrule
\end{tabular}
\end{table}

\begin{table}[!t]
\centering
\caption{Ablation of online fast-weight updates and cross-observation memory persistence on LIBERO-Plus. Under a shared SReWAM checkpoint, online updates improve success by 0.81 percentage points, while retaining fast weights across observations provides an additional 0.97-point gain. All configurations use one-step inference; the Static Adapter baseline is trained separately.
}

\label{tab:memory_ablation}
\small
\setlength{\tabcolsep}{4pt}
\renewcommand{\arraystretch}{1.0}
\begin{tabular*}{\columnwidth}{@{}l@{\extracolsep{\fill}}ccc@{}}
\toprule
\textbf{Method}
& \textbf{Online}
& \textbf{Persistent}
& \textbf{Avg. (\%)} \\
\midrule
Static Adapter
& No & No & 77.20 \\
SReWAM (Slow Only)
& No & No & 77.12 \\
SReWAM (Reset)
& Yes & No & 77.93 \\
\rowcolor[gray]{0.94}
\textbf{SReWAM (Ours)}
& Yes & Yes & \textbf{78.9} \\
\bottomrule
\end{tabular*}
\end{table}

\begin{table}[t]
\centering
\caption{Real-world can-to-plate manipulation under the original and white-tabletop settings. SReWAM (WE) achieves 100\% success in both settings. Representative initial observations, successful SReWAM outcomes, and $\pi_{0.5}$ failures are shown from two camera views. All methods use one-step inference.}
\label{tab:real_robot}
\scriptsize
\renewcommand{\arraystretch}{1.0}
{\centering
\textbf{Task: Pick up the can and place it on the plate}
\par}
\vspace{3pt}
\newcommand{\robotimg}[1]{%
  \raisebox{-0.5\height}{%
    \includegraphics[width=0.132\columnwidth]{#1}}%
}
\newcommand{\robotblank}{%
  \raisebox{-0.5\height}{%
    \begingroup
    \setlength{\fboxsep}{0pt}%
    \framebox[0.132\columnwidth]{%
      \rule{0pt}{0.132\columnwidth}%
    }%
    \endgroup
  }%
}
\newcommand{\robotfail}[1]{%
  \IfFileExists{#1}{%
    \robotimg{#1}%
  }{%
    \robotblank
  }%
}
\begin{tabular}{
@{}l
@{\hspace{2pt}}c@{\hspace{1pt}}c
@{\hspace{5pt}}c@{\hspace{1pt}}c
@{\hspace{5pt}}c@{\hspace{1pt}}c
@{}
}
& \multicolumn{2}{c}{\textbf{Initial Setup}}
& \multicolumn{2}{c}{\textbf{Task Completed}}
& \multicolumn{2}{c}{\textbf{Task Failure}} \\[2pt]
\textbf{Original}
& \robotimg{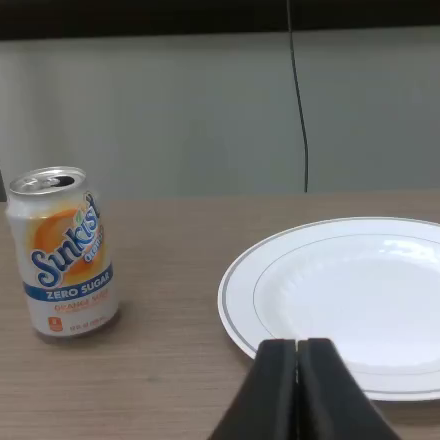}
& \robotimg{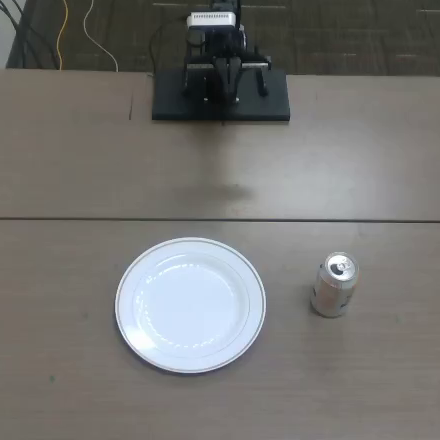}
& \robotimg{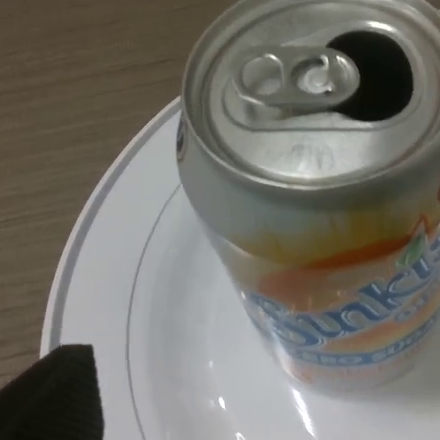}
& \robotimg{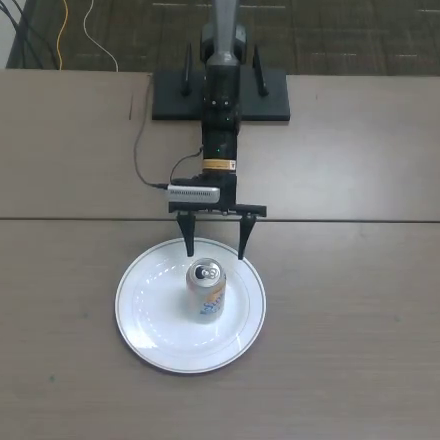}
& \robotfail{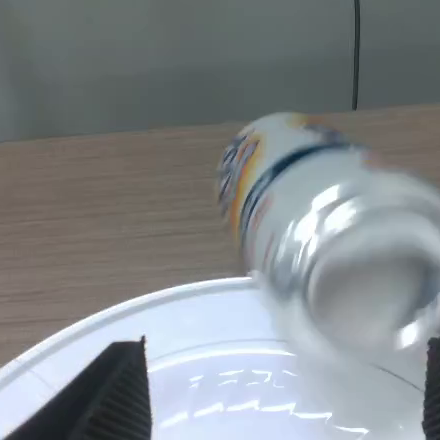}
& \robotfail{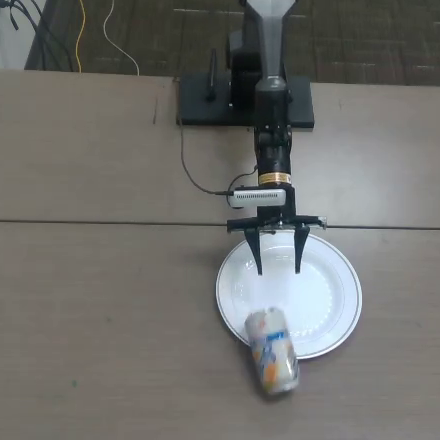}
\\[2pt]
\textbf{White}
& \robotimg{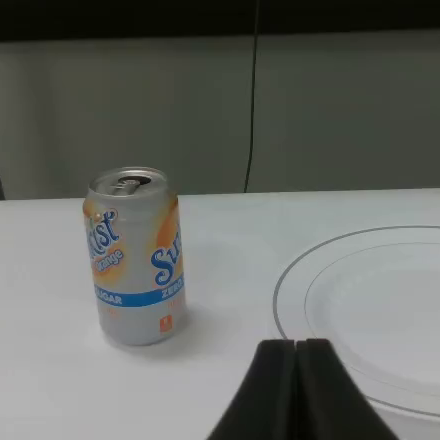}
& \robotimg{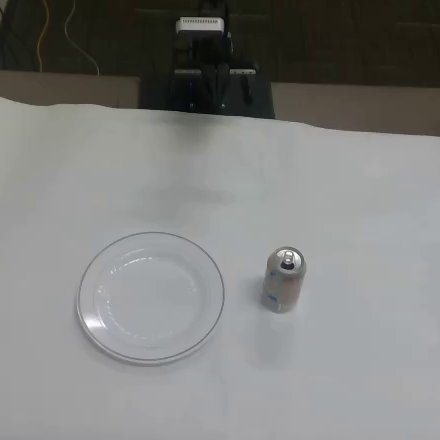}
& \robotimg{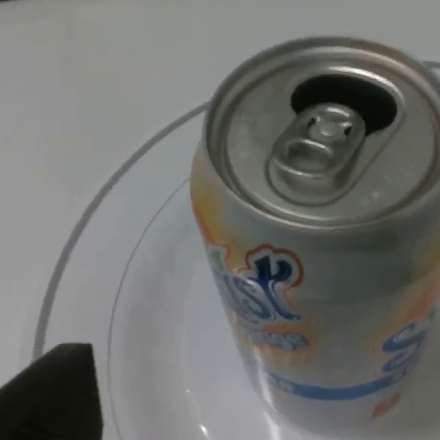}
& \robotimg{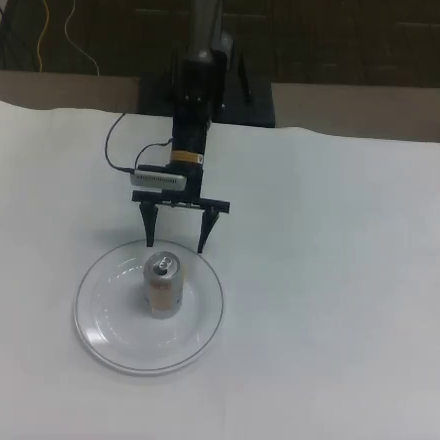}
& \robotfail{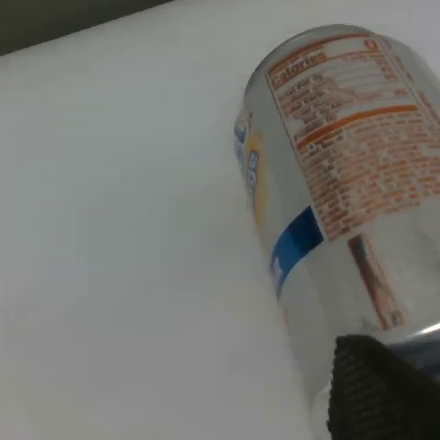}
& \robotfail{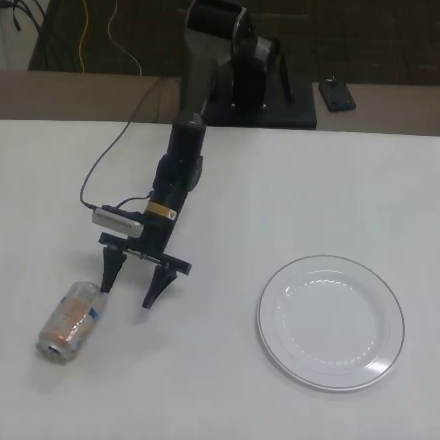}
\\
\end{tabular}
\vspace{5pt}
\setlength{\tabcolsep}{3pt}
\begin{tabular*}{\columnwidth}{
@{}l@{\extracolsep{\fill}}cc@{}
}
\toprule
\textbf{Method}
& \textbf{Original (\%)} $\uparrow$
& \textbf{White (\%)} $\uparrow$ \\
\midrule
$\pi_{0.5}$
& 83.3
& 40.0 \\
$\pi_{0.5}$ + JEPA Obj.
& 93.3
& 80.0 \\
\rowcolor[gray]{0.94}
\textbf{$\pi_{0.5}$ + SReWAM}
& \textbf{100.0}
& \textbf{100.0} \\
\bottomrule
\end{tabular*}
\end{table}

\subsection{Real-World Evaluation}

\paragraph{Robot Setup and Evaluation.}
We evaluate SReWAM on an AgileX PiPER 6-DoF robotic arm equipped with two Intel RealSense D435i RGB cameras (front and wrist views, $640 \times 480$ at 30 FPS). Policy inference and online memory updates run on an NVIDIA RTX A6000 GPU. The task requires picking up a can and placing it on a plate. We evaluate the original tabletop and a held-out white-tabletop setting, which introduces an out-of-distribution (OOD) appearance shift absent from the training demonstrations. Unlike the frozen-backbone simulation experiments, all three real-world policies are fine-tuned from the official $\pi_{0.5}$ base checkpoint using the same demonstration dataset and optimization settings for 2,000 optimizer updates, with model-specific training objectives and components. All methods use one-step inference, predicting 16 actions and executing the first eight before replanning. Each method is evaluated over 30 trials in the original setting and 10 trials in the OOD setting.

\noindent\textbf{Results.}
As shown in Table~7, SReWAM achieves 100\% success in both the original and white-tabletop settings, compared with 83.3\%/40.0\% for $\pi_{0.5}$ and 93.3\%/80.0\% for $\pi_{0.5}$ + JEPA Obj. Relative to the JEPA-adapted baseline, SReWAM improves success by 6.7 and 20.0 percentage points, respectively. These gains correspond to two additional successful trials in each setting, with all methods using one-step action generation. The larger percentage-point improvement under the held-out tabletop appearance suggests that memory-augmented conditioning may help maintain task performance under this visual shift. However, the limited number of trials, particularly in the white-tabletop setting, does not establish general robustness across unseen environments.

\section{Conclusion and Limitations}

We introduced SReWAM, a sparse memory framework for one-step world-action models that constructs persistent fast-weight memory from pretrained JEPA representations. Its observation-aligned write-then-read mechanism separates visual memory writing from policy-conditioned retrieval, providing historical conditioning through selected action-expert layers. On LIBERO-Plus, SReWAM improves one-step success by 1.8 and 0.7 percentage points across two frozen JEPA-based backbones, while achieving comparable performance to dense memory with 75\% fewer additional trainable parameters. Shared-checkpoint ablations isolate the effects of online adaptation and cross-observation persistence, with improvements of 0.81 and 0.97 percentage points, respectively. Counterfactual memory interventions further indicate that performance is sensitive to the specific historical fast-weight content, providing complementary evidence for the role of accumulated memory in one-step action generation.

Despite these improvements, online fast-weight updates introduce additional inference latency compared with the original one-step backbone. Performance may also depend on the quality of pretrained visual representations, and the training context spans at most 128 frames. Our RoboTwin evaluation covers only four selected Clean-setting tasks, while real-world experiments are restricted to a single manipulation task and two tabletop appearances. Moreover, the real-robot variants differ in their training objectives, so the observed improvements cannot be attributed solely to persistent memory. Future work will investigate longer-term memory, more efficient online updates, and broader evaluations across manipulation tasks and environmental shifts.

\clearpage


{
    \small
    \bibliographystyle{ieeenat_fullname}
    \bibliography{main}
}

\clearpage
\setcounter{page}{1}
\setcounter{section}{0}
\maketitlesupplementary




\section{Simulation Implementation Details}
\label{app:simulation}

\subsection{JEPA-WAM Variants on LIBERO}
\label{app:jepa_libero}

For Part I, we use JEPA-WAM~\cite{lin2026jepa} as the backbone. We train four TTT memory variants on LIBERO and evaluate them on LIBERO-Plus, as summarized in Table~\ref{tab:ttt_variants}. Wrapper variants (WE and WI) insert TTT modules at four selected blocks of the 16-block DiT action expert, while Inline variants (IE and II) use all 16 blocks. Explicit variants (WE and IE) construct memory keys and values from current V-JEPA representations; Implicit variants (WI and II) instead use the complete post-attention DiT token sequence, including state, register, future latent, and action tokens. All four variants retrieve memory through state, register, and action tokens.

The pretrained JEPA-WAM backbone remains frozen. We train on windows of 16 observations sampled at a stride of eight frames, using eight-observation truncated backpropagation through time (TBPTT) and a global batch size of 64. Optimization uses AdamW with a peak learning rate of $2\times10^{-5}$, weight decay of $10^{-5}$, and a configured maximum of 30,000 optimizer steps. Checkpoints are saved periodically and selected using validation action-prediction loss rather than rollout success.

\begin{table}[t]
\centering
\caption{TTT memory variants for JEPA-WAM trained on LIBERO and evaluated on LIBERO-Plus. All variants use the same memory retrieval token types. DiT block positions are one-based.}
\label{tab:ttt_variants}
\small
\setlength{\tabcolsep}{4pt}
\renewcommand{\arraystretch}{1.1}
\begin{tabular}{@{}llll@{}}
\toprule
\textbf{Variant} & \textbf{Integration} & \textbf{K/V Source} & \textbf{DiT Blocks} \\
\midrule
WE & Wrapper & Current V-JEPA & 4, 8, 12, 16 \\
IE & Inline & Current V-JEPA & All 16 \\
WI & Wrapper & Post-attn. DiT tokens & 4, 8, 12, 16 \\
II & Inline & Post-attn. DiT tokens & All 16 \\
\bottomrule
\end{tabular}
\end{table}

\begin{figure*}[t]
    \centering
    \includegraphics[
        width=\textwidth,
        trim=0.1cm 13cm 0.1cm 0.5cm,
        clip
    ]{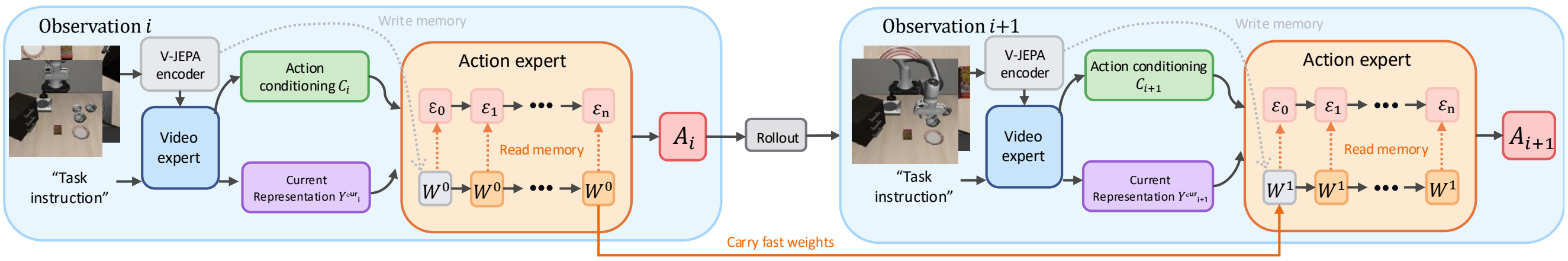}
    \caption{Four-step inference with observation-aligned persistent memory. At each observation, current JEPA representations update the fast-weight memory once before action generation. The updated memory remains fixed and is repeatedly queried across denoising steps. Here, $\epsilon_0,\epsilon_1,\ldots,\epsilon_n$ schematically denote successive denoising states, while $W^0$ and $W^1$ represent the effective memory states after updates at consecutive observations. The final memory state is carried forward and updated upon receiving the next observation.}
    \label{fig:four_step_inference}
\end{figure*}
\subsection{TTT Memory Implementation Details}
\label{app:ttt_implementation}

\paragraph{Action Expert Token Construction and Conditioning.}
At each observation, the pretrained vision-language backbone supplies action-conditioning tokens $C_i$ from its final hidden states at action-placeholder positions. These tokens remain separate from the DiT policy-token sequence and provide keys and values for cross-attention, with DiT tokens serving as queries.

In SReWAM, the DiT sequence comprises one MLP-encoded proprioceptive state token, introduced learnable register tokens, learned future latent tokens, and action tokens encoded from the noisy action chunk and flow timestep. The timestep additionally modulates the DiT through a separate time-conditioning pathway. The DiT processes these tokens through interleaved self-attention and cross-attention, while only the outputs at action-token positions are decoded into predicted flow velocities.

Register tokens provide additional learnable representation slots within the DiT sequence and participate in policy-conditioned memory retrieval. Future latent tokens are learned internal placeholders rather than representations extracted from future observations. In the Wrapper-Explicit (WE) configuration, TTT queries and residual injection are restricted to post-attention state, register, and action tokens, excluding future latent tokens. Memory-writing keys and values are constructed from current V-JEPA representations rather than action-conditioning tokens or predicted future features. Neither register nor future latent tokens maintain persistent memory across observations; this information is carried by the online-updated TTT fast weights.

\paragraph{Memory Representations.}
For JEPA-WAM, Explicit memory uses 1024-dimensional current V-JEPA features. The features are L2-normalized; keys additionally receive RMSNorm, while values are used without learned projections. Queries are projected from selected post-attention DiT tokens in the 1536-dimensional action-expert space to the 1024-dimensional memory space. Implicit memory uses the complete post-attention DiT token sequence for keys and values, with memory dimension 1536. Queries and keys receive temporal rotary positional encoding (RoPE), whereas values do not. Each TTT module maintains a separate temporal-position counter for each sample, initialized to zero at episode boundaries and incremented once per valid fast-weight update rather than per memory token. All key and query tokens associated with the same update use the same RoPE position.

For the JEPA-adapted $\pi_{0.5}$ backbone, Explicit memory uses 1408-dimensional current V-JEPA features, with policy queries projected from the 1024-dimensional action-expert space. Each enabled TTT module has an independent two-layer GELU MLP with hidden width $2d_m$, where $d_m$ is the memory dimension. The MLP dimensions are $1024\rightarrow2048\rightarrow1024$ for JEPA-WAM Explicit, $1536\rightarrow3072\rightarrow1536$ for JEPA-WAM Implicit, and $1408\rightarrow2816\rightarrow1408$ for $\pi_{0.5}$ Explicit.

\paragraph{Gating and Initialization.}
The online update gate produces one scalar per sample and enabled layer. For JEPA-WAM, selected post-attention tokens are mean-pooled and RMS-normalized before a two-layer SiLU--sigmoid MLP (1536--3072--1). For $\pi_{0.5}$, tokens are RMS-normalized before mean pooling and processed by a 1024--2048--1 MLP with the same activations. The inner step size is parameterized by $\operatorname{softplus}(\alpha_\ell)$, with $\alpha_\ell$ initialized to $0.01$ (an initial effective step size of approximately $0.698$). Retrieved features are scaled channel-wise by $G^{(\ell)}=\tanh(a_\ell)$, with $a_\ell$ initialized from a zero-mean Gaussian of standard deviation $10^{-4}$. The residual scale has width 1536 for JEPA-WAM and 1024 for $\pi_{0.5}$. Fast-weight offsets are initialized to zero.

\paragraph{Online Memory Optimization.}
The inner self-supervised memory loss is an MSE averaged over memory tokens and feature dimensions. Under the inference schedule used in our experiments, each valid observation triggers one gated fast-weight gradient update per enabled module, followed by retrieval from the updated memory MLP. The outer flow-matching objective optimizes trainable memory-related slow parameters through differentiable inner updates. During each eight-observation TBPTT segment, gradients propagate through the updates; at segment boundaries, fast-weight values are retained while their computation graphs are detached.

\subsection{Static Adapter Baseline}
\label{sec:static_adapter}

We train a separate Static Adapter baseline to compare persistent fast-weight memory with current-observation feature conditioning. Starting from the pretrained JEPA-WAM checkpoint, we freeze the perception backbone and original action expert and optimize only the added wrapper adapter and register embeddings. The adapter uses current V-JEPA representations and policy-token queries, injecting features through learned MLP transformations, gating, and residual connections. Unlike SReWAM, it processes observations independently, without online fast-weight gradients or persistent memory.

Training uses the four LIBERO suites after removing empty-action samples, with a single observation per window and no cross-observation state. We optimize the action flow-matching objective with future visual alignment disabled, using four NVIDIA H100 GPUs, a global batch size of 32 (eight per GPU), AdamW with learning rate $2\times10^{-5}$ and weight decay $10^{-5}$, and a maximum budget of 30,000 steps. The selected best-validation checkpoint was saved at step 20,000; this is the evaluated checkpoint, not the maximum training step.

\paragraph{Memory Ablation Protocol.}
All configurations in Table~\ref{tab:memory_ablation} use one-step inference. Static Adapter is trained separately, whereas Slow Only, Reset, and Full SReWAM use the same trained Wrapper-Explicit (WE) checkpoint. Slow Only disables online fast-weight updates. Reset performs an update for each observation but does not carry the resulting fast-weight offsets into the next observation. Full SReWAM carries updated offsets across observations and resets them at episode boundaries. Static Adapter and Slow Only achieve 77.20\% and 77.12\%, respectively; Reset achieves 77.93\%, and Full SReWAM achieves 78.90\%. Among the shared-checkpoint variants, observation-level updating improves success by 0.81 percentage points (Slow Only to Reset), while carrying fast weights across observations provides a further 0.97-point gain (Reset to Full). Since Static Adapter is trained separately, the shared-checkpoint comparisons offer the more direct evidence for the contributions of online adaptation and persistence.

\subsection{Wrapper-Explicit Implementation on $\pi_{0.5}$}
\label{app:pi05_we}

For Part II, we use the JEPA-adapted $\pi_{0.5}$ backbone~\cite{intelligence2025pi_,lin2026jepa} with Wrapper-Explicit (WE) memory. Its 18-block action expert has TTT wrappers at blocks 4, 8, 12, and 16 (one-based), after the attention residual and before the feed-forward operation. These positions are predetermined rather than chosen by a layer-placement search.

Memory writing uses current image-patch representations from the frozen V-JEPA 2.1 encoder. The TTT queries and gated residuals involve the state token, 16 introduced register tokens, and action tokens; the 32 learned future latent tokens receive no direct memory residuals. Action-conditioning tokens are supplied separately through the pretrained backbone's cross-attention pathway and are not used as WE memory-writing keys or values.

Training updates TTT-related parameters and the introduced register embeddings through the outer flow-matching objective, while the pretrained vision-language backbone, action expert, state projection, V-JEPA encoder, JEPA predictor, and future-token embeddings remain frozen. No auxiliary future-pair JEPA alignment loss is applied. Training windows remain within individual episodes, invalid observations are masked, and each independent window starts with zero fast-weight offsets.

\subsection{$\pi_{0.5}$ Training on LIBERO}
\label{app:pi05_libero}

For LIBERO, we train WE with an action horizon of 10 and a global batch size of 16, initialized from the released JEPA-adapted $\pi_{0.5}$ LIBERO checkpoint. Explicit memory writing uses current V-JEPA representations from the agent-view camera, while the policy also receives wrist-camera observations. Each training window contains 16 observations sampled at stride eight, with a TBPTT length of eight updates.

The JEPA-adapted $\pi_{0.5}$ backbone retains its flow-matching convention, distinct from JEPA-WAM. With $u_i\sim\operatorname{Beta}(1.5,1)$, we use
\begin{equation}
\begin{aligned}
    t_i &= 0.999u_i+0.001,\\
    \mathbf{x}_i^{t_i} &= t_i\boldsymbol{\epsilon}_i+(1-t_i)\mathbf{a}_i,\\
    \mathbf{v}_i^\star &= \boldsymbol{\epsilon}_i-\mathbf{a}_i,
\end{aligned}
\end{equation}
where $\boldsymbol{\epsilon}_i$ is Gaussian noise and $\mathbf{a}_i$ is the ground-truth action chunk. The same convention is used for RoboTwin.

\subsection{$\pi_{0.5}$ Training on RoboTwin}
\label{app:pi05_robotwin}

For RoboTwin 2.0, we train SReWAM (WE) on four tasks selected from RoboTwin Clean-20: \texttt{adjust\_bottle}, \texttt{dump\_bin\_bigbin}, \texttt{shake\_bottle}, and \texttt{press\_stapler}. We use 50 training episodes per task, totaling 200 episodes.

The policy receives three camera views (head and two wrist cameras), while Explicit memory writing uses current V-JEPA representations from the head-camera view. Training uses an action horizon of 50, windows of 16 observations sampled at a stride of eight, a TBPTT length of eight updates, and a global batch size of 16.

We initialize SReWAM from the released JEPA-adapted $\pi_{0.5}$ Clean-20 checkpoint rather than a previously TTT-trained model. Training episodes are selected by episode indices from the larger Clean-20 collection. Evaluation is conducted on the same four tasks under the Clean setting.

\subsection{Inference-Time Memory Update Schedules}
\label{app:inference_schedule}

The update-and-retrieval mechanism is described in Sec.~\ref{sec:persistent_action}. All reported multi-step experiments use an observation-aligned first-denoise schedule: each valid new observation triggers one online fast-weight update per enabled TTT module during the first denoising evaluation. The updated memory is immediately available for retrieval and remains fixed during subsequent denoising steps. Fast-weight offsets and temporal-position counters persist across valid observations within an episode and reset at episode boundaries.

For one-step inference, memory is updated and queried within a single action-expert evaluation. For four-step inference, memory is updated only during the first denoising evaluation and reused during the remaining three steps, as illustrated in Fig.~\ref{fig:four_step_inference}. During the four denoising evaluations, retrieval queries use the same RoPE position for the current observation; the temporal-position counter advances only at the next valid memory update. In contrast, the original JEPA-WAM baseline has no TTT memory and performs standard one-step or four-step action generation without online fast-weight updates.

\subsection{Evaluation of Author-Released Checkpoints}
\label{sec:released_checkpoint_eval}

We independently evaluate three checkpoints released by the JEPA-WAM authors without retraining: JEPA-WAM on LIBERO, JEPA-adapted $\pi_{0.5}$ on LIBERO, and JEPA-adapted $\pi_{0.5}$ on RoboTwin.

\paragraph{LIBERO-Plus.}
We evaluate the released JEPA-WAM checkpoint and the JEPA-adapted $\pi_{0.5}$ LIBERO checkpoint at step 29,999. Both models are evaluated under one-step and four-step denoising on the Spatial, Object, Goal, and LIBERO-10 suites, covering 10,030 perturbed evaluation rollouts per configuration. The policy executes eight predicted actions before replanning. Reported results are those obtained by our evaluation pipeline and are distinct from the original paper's reported JEPA-WAM average.

\paragraph{Comparison with Published Results.}
Both JEPA-based baselines are independently evaluated using the authors' publicly released checkpoints. The original JEPA-WAM reports 79.2\% success on LIBERO-Plus, whereas our four-step evaluation obtains 77.8\%. For the JEPA-adapted $\pi_{0.5}$, the published result is 86.3\%, compared with 86.0\% under our four-step evaluation. Differences in evaluation settings may contribute to these discrepancies, but we do not attribute them to a specific cause. All SReWAM improvements are computed against the corresponding locally evaluated baselines under matched denoising budgets and the same evaluation protocol.

\paragraph{RoboTwin 2.0.}
For our local evaluation of $\pi_{0.5}$ + JEPA Obj., we use the publicly released JEPA-adapted $\pi_{0.5}$ Clean-20 checkpoint at step 14,999 without retraining. We evaluate four tasks under the Clean setting: Adjust Bottle, Dump Bin Bigbin, Shake Bottle, and Press Stapler. Each task uses 100 expert-screened evaluation episodes, with replanning after executing eight actions from each predicted 50-action chunk.

The JEPA-adapted baseline is evaluated with one-step and four-step denoising, while SReWAM (WE) is evaluated with one-step denoising. Reported averages are computed over the four tasks. Published baseline results marked with $^\dagger$ in Table~\ref{tab:robotwin} are taken from JEPA-WAM~\cite{lin2026jepa} and are distinguished from our local evaluations. The published $\pi_{0.5}$ baseline uses 10-step denoising.

Notably, the released JEPA-adapted $\pi_{0.5}$ checkpoint achieves 1\% success on Press Stapler with four-step denoising, compared with 40\% under one-step inference in our evaluation. We reproduced the 1\% four-step result on both NVIDIA RTX A6000 and H100 GPUs using the same released checkpoint and evaluation pipeline. We report this discrepancy without attributing it to a specific cause.

\subsection{Inference Latency Measurement Protocol}
\label{app:latency_protocol}

We measure inference latency using preloaded LIBERO-Plus Spatial observations on a single NVIDIA H100 PCIe GPU. The JEPA-WAM/WE benchmark and the WE/IE one-step comparison are conducted in separate runs. Within each run, configurations are evaluated sequentially using the same GPU and observation sequence. For the WE/IE comparison, both methods process 230 observations from 10 episodes, with fast-weight memory reset at episode boundaries. Both use one denoising step and one online memory update per observation, followed by retrieval from the updated memory.

Each configuration uses 30 warmup calls followed by 200 timed calls, with CUDA synchronization before and after each measurement. Action-expert latency includes action-head inference, online memory updates, and memory retrieval. End-to-end (E2E) latency additionally includes input preprocessing, backbone computation, and action postprocessing, while excluding observation acquisition, environment execution, model loading, and episode resets.

We report the arithmetic mean latency per predicted action chunk. As shown in Table~\ref{tab:inference_latency}, one-step WE achieves 30.86 ms action-expert latency and 163.79 ms E2E latency, compared with 70.72 ms and 175.02 ms for IE, respectively. WE integrates four TTT-equipped DiT blocks, whereas IE uses 16.

\subsection{Statistical Analysis of LIBERO-Plus Results}
\label{sec:supp_statistical_analysis}

\paragraph{Paired Bootstrap Protocol.}
We estimate the uncertainty of success-rate differences using episode-level binary outcomes from the 10,030 LIBERO-Plus evaluation episodes per configuration across four suites. For JEPA-WAM comparisons, episodes are paired by suite, task ID, and episode index; for the JEPA-adapted $\pi_{0.5}$ comparisons, recorded episode seeds are additionally matched. We draw 5,000 paired bootstrap resamples with replacement within each suite--perturbation-category stratum, using the same sampled episode indices for both configurations in a comparison. In each resample, we pool outcomes across the four suites within each perturbation category, compute the success-rate difference for that category, and then take the unweighted mean of the seven category-wise differences, matching the main-paper metric. The 2.5th and 97.5th percentiles of the bootstrap differences form the reported 95\% confidence interval. Table~\ref{tab:paired_bootstrap} summarizes the three comparisons.

\begin{table}[t]
\centering
\caption{
Paired bootstrap analysis on LIBERO-Plus.
SReWAM (WE) improves one-step success on both
JEPA-based backbones, while retaining fast-weight
memory across observations provides an additional
benefit over resetting it.
All comparisons use one-step inference.
Full and Reset share the same SReWAM checkpoint.
Differences and 95\% confidence intervals are
reported in percentage points.
}
\label{tab:paired_bootstrap}
\footnotesize
\setlength{\tabcolsep}{2pt}
\renewcommand{\arraystretch}{1.12}
\begin{tabular*}{\columnwidth}{
@{}p{0.58\columnwidth}
@{\extracolsep{\fill}}rr@{}
}
\toprule
\textbf{Comparison}
& \textbf{Difference}
& \textbf{95\% CI} \\
\midrule
SReWAM vs. JEPA-WAM
& $+1.80$ & $[+1.16,\,+1.91]$ \\
$\pi_{0.5}$ + SReWAM vs. $\pi_{0.5}$ + JEPA Obj.
& $+0.70$ & $[+0.46,\,+0.89]$ \\
SReWAM: Full vs. Reset
& $+0.97$ & $[+0.76,\,+1.37]$ \\
\bottomrule
\end{tabular*}
\end{table}

\subsection{Counterfactual Memory Intervention}
\label{sec:counterfactual_memory}

\paragraph{Experimental Protocol.}
To examine whether historical fast-weight content influences action generation, we conduct a counterfactual memory intervention using the trained JEPA-WAM SReWAM Wrapper-Explicit (WE) checkpoint with one-step inference. After four observation-level memory updates, each rollout is branched from an identical simulator state into three conditions: Real History retains the accumulated fast weights, Cleared History zeros the historical fast-weight offsets, and Swapped History replaces them with offsets from another episode of the same task with a matched update count.

At the branching point, the current observation, task instruction, proprioception, action-sampling noise, and temporal-position counters are held fixed. Only the historical fast-weight offsets are modified, and all branches follow the same write-then-read update schedule before independently continuing their rollouts. Simulator-state restoration and action-sampling consistency are checked using identical-history control rollouts.

We preselect 1,000 target episodes across four LIBERO suites and seven perturbation categories. Of these, 971 yield valid matched evaluations under all three memory conditions, comprising 240 Spatial, 243 Object, 246 Goal, and 242 LIBERO-10 episodes. The remaining 29 episodes are excluded due to mismatches between replayed and reference observations (15 agent-view and 14 wrist-view mismatches). These episodes are not counted as task failures, and no replacement episodes are introduced. All comparisons use the same 971 matched target episodes. Within each suite, success rates are averaged equally across the seven perturbation categories, and the overall average is the unweighted mean across the four suites. 
The counterfactual success rate is
macro-averaged across perturbation
categories and task suites:

\begin{equation}
\widehat{\mathrm{SR}}_{\mathrm{macro}}
=
\frac{100}{4}
\sum_{s=1}^{4}
\frac{1}{7}
\sum_{c=1}^{7}
\frac{1}{n_{s,c}}
\sum_{j=1}^{n_{s,c}}
z_{s,c,j},
\label{eq:counterfactual_macro}
\end{equation}

where $z_{s,c,j}\in\{0,1\}$ denotes
episode success and $n_{s,c}$ is the
number of valid matched episodes in
suite $s$ and perturbation category $c$.
The same matched episodes and aggregation
are used for all three memory conditions.

\begin{table}[t]
\centering
\caption{Counterfactual memory intervention on LIBERO-Plus.
Success rates (\%) are reported across four task suites.
Real retains the original historical fast weights,
Cleared removes historical offsets, and Swapped uses
memory from another episode of the same task.
Success rates are macro-averaged across seven perturbation
categories within each suite and then averaged equally
across the four suites.}
\label{tab:counterfactual_memory}
\footnotesize
\setlength{\tabcolsep}{4pt}
\renewcommand{\arraystretch}{1.12}
\begin{tabular}{@{}lrrr@{}}
\toprule
LIBERO Suite & Real & Swapped & Cleared \\
\midrule
Spatial
    & \textbf{83.16} & 83.15 & 82.76 \\
Object
    & \textbf{84.61} & 83.39 & 84.18 \\
Goal
    & \textbf{79.03} & 74.22 & 76.61 \\
LIBERO-10 
    & \textbf{66.33} & 65.15 & 65.75 \\
\midrule
Average
    & \textbf{78.28} & 76.48 & 77.33 \\
\bottomrule
\end{tabular}
\end{table}

\paragraph{Results and Discussion.}
As shown in Table~\ref{tab:counterfactual_memory}, Real History achieves the highest average success rate (78.28\%), compared with Cleared History (77.33\%) and Swapped History (76.48\%), corresponding to improvements of 0.95 and 1.80 percentage points, respectively. Real History numerically outperforms both alternative memory conditions across all four suites. The largest Real-to-Swapped advantage occurs on Goal tasks, where success increases from 74.22\% to 79.03\% (+4.81 points).

These results suggest that downstream manipulation performance is sensitive to the specific historical fast-weight state rather than merely the presence of a memory pathway. Unlike the full-episode persistence ablation, this intervention changes memory at an intermediate observation while holding the current policy inputs fixed. The results provide exploratory evidence of memory-content sensitivity, without establishing statistical significance. Together with the shared-checkpoint persistence ablation in Table~\ref{tab:memory_ablation}, these findings provide complementary evidence for the role of historical memory in one-step action generation.

\section{Real-World Training and Evaluation Details}
\label{app:real_world_details}

\subsection{Demonstrations and Data Split}
\label{app:real_robot_data}

We collect a single-task demonstration dataset using an AgileX PiPER 6-DoF robotic arm for the instruction ``Pick up the can and place it on the plate.'' The dataset is converted to LeRobot v2.1 format and contains 150 episodes (19,246 recorded frames at 30\, Hz). Each sample includes synchronized front-view and wrist-view RGB images, robot proprioceptive observations, and demonstrated actions. Both cameras are Intel RealSense D435i devices, recording $640\times480$ RGB images (width $\times$ height); neither depth measurements nor IMU signals are used for policy learning.

We reserve eight complete episodes (IDs 55, 76, 108, 109, 124, 127, 137, and 148) for offline validation and use the remaining 142 episodes for training. Data normalization is computed using only the training split. The validation set consists of temporal windows sampled from held-out episodes, with no overlap of episodes between training and validation. All three real-robot policies---$\pi_{0.5}$, $\pi_{0.5}$ + JEPA Obj., and $\pi_{0.5}$ + SReWAM (WE)---use the same demonstrations and episode-level split.

\subsection{Observation and Action Representations}
\label{app:real_robot_representation}

The policy receives two RGB views and a 14-dimensional robot state. The state comprises six arm joint positions, one gripper position, a six-dimensional end-effector pose, and one scalar gripper force/torque feedback value.

The physical robot action contains six joint-position commands and one gripper command. During training, the joint commands are represented relative to the current observed joint positions:
\begin{equation}
\mathbf{a}_{t,h} =
\begin{bmatrix}
\mathbf{q}^{\mathrm{cmd}}_{t,h}
-\mathbf{q}_t \\
g^{\mathrm{cmd}}_{t,h}
\end{bmatrix}
\in\mathbb{R}^{7},
\label{eq:real_robot_action}
\end{equation}
where $\mathbf{q}_t\in\mathbb{R}^{6}$ denotes the current joint state and $g^{\mathrm{cmd}}_{t,h}$ is an absolute gripper command. The action horizon is $h=0,\ldots,15$. Any padding required by the internal $\pi_{0.5}$ action representation is discarded when decoding the corresponding seven-dimensional robot command.

The front and wrist cameras are mapped to the $\pi_{0.5}$ base-camera and left-wrist inputs, respectively. Since the robot has no right-wrist camera, that input is masked. The policy predicts action chunks of length $H=16$. The image acquisition rate of 30\, Hz describes the recorded RGB streams and should not be interpreted as the robot control rate.

\subsection{Common Fine-Tuning Protocol and Model Variants}
\label{app:real_robot_variants}

We compare three variants initialized from the official $\pi_{0.5}$ base policy: (i) $\pi_{0.5}$, (ii) $\pi_{0.5}$ + JEPA Obj., and (iii) $\pi_{0.5}$ + SReWAM (WE). All variants use the same real-robot training dataset, episode split, optimizer configuration, and training budget of 2,000 optimizer updates. Their model-specific components and objectives differ by design. The $\pi_{0.5}$ baseline is supervised by the action flow-matching objective. The JEPA-adapted variant additionally employs joint current--future representation alignment, while SReWAM further incorporates explicit representation-driven TTT memory. Thus, the comparison holds the demonstrations and overall fine-tuning budget fixed, but does not equate the number of trainable parameters or the optimization objective across variants.

Unlike the frozen-backbone simulation adaptation, real-robot SReWAM follows a joint fine-tuning protocol: the policy backbone, action expert, JEPA prediction components, and added TTT wrappers are optimized together. The V-JEPA teacher remains frozen. This real-world protocol is separate from the frozen-backbone experiments on LIBERO-Plus and RoboTwin 2.0 described in the main paper.

\subsection{JEPA Supervision and Persistent TTT Memory}
\label{app:real_robot_jepa_ttt}

\paragraph{Training-time JEPA Objective.}
We use a frozen V-JEPA 2.1 ViT-g/384 encoder $E_J$ to construct joint current--future visual targets. Given the front-camera image $I_t$ at time $t$, we select a future frame at $t'=\min(t+31, T_e)$, where $T_e$ denotes the final valid frame index of the episode. The two frames are stacked along the temporal dimension and jointly encoded:

\begin{equation}
\mathbf{Y}_{t,t'} = \operatorname{sg}\left[E_J\left( \operatorname{Stack}_{\mathrm{time}}(I_t,I_{t'})\right)\right],
\label{eq:real_jepa_target}
\end{equation}

where $\mathbf{Y}_{t,t'}\in \mathbb{R}^{576\times1408}$ and $\operatorname{sg}$ denotes stop-gradient. The predicted joint representation $\hat{\mathbf{Y}}_t$ is supervised by the mean patch-wise cosine distance:

\begin{equation}
\mathcal{L}_{\mathrm{JEPA}} =
\frac{1}{BN}
\sum_{b=1}^{B}\sum_{p=1}^{N}
\left[
1-\cos\left(
\hat{\mathbf{y}}_{b,p},
\mathbf{y}_{b,p}
\right)
\right],
\label{eq:real_jepa_loss}
\end{equation}

where $N=576$ is the number of spatial patch tokens. The teacher targets are precomputed and cached for training. Separately encoded current-frame V-JEPA features are used for TTT memory writing. Future observations and cached transition targets are not required during deployment.

\paragraph{Observation-aligned fast-weight memory.}
SReWAM uses the Wrapper-Explicit (WE) configuration, with TTT modules at action-expert block indices $\{3,7,11,15\}$ in zero-based indexing ($\{4,8,12,16\}$ in the one-based notation of the main paper). Current V-JEPA visual representations construct memory-writing keys and values, while post-attention policy representations query the updated memory. The temporal training context consists of 16 observations sampled every eight recorded frames, with truncated backpropagation through time (TBPTT) over eight observation-level updates.

The real-robot implementation follows the same initialization, persistence, and reset rules as the main method. Fast-weight offsets are initialized to zero at the start of each independent training window or evaluation episode. Each valid observation triggers one gated self-supervised write per enabled TTT module, followed by retrieval from the updated memory for action generation. Fast-weight values are preserved across successive observations and across TBPTT segment boundaries; computation graphs are detached at segment boundaries. At the end of an episode, fast-weight offsets and temporal-position counters are reset. Under multi-step denoising, the write is performed only during the first denoising evaluation of each observation, whereas the real-robot results here use one-step inference.

\paragraph{Joint Training Objective.}
For JEPA-supervised configurations, the outer
optimization objective combines action flow matching
with joint current--future representation alignment:

\begin{equation}
\begin{aligned}
\mathcal{L}_{\mathrm{real}}(u)
    &= \mathcal{L}_{\mathrm{FM}}
      + \lambda_{\mathrm{JEPA}}(u)\mathcal{L}_{\mathrm{JEPA}},\\
\lambda_{\mathrm{JEPA}}(u)
    &= 0.1\min\!\left(1,\frac{u}{500}\right),
\end{aligned}
\label{eq:real_robot_joint_training}
\end{equation}

where $u$ counts optimizer updates.
The JEPA loss weight increases linearly from zero
to $0.1$ over the first 500 updates.
For the original $\pi_{0.5}$ baseline,
$\lambda_{\mathrm{JEPA}}=0$.
The inner TTT objective, when enabled, is the
self-supervised memory regression loss used for
online fast-weight updates; it is distinct from
the outer JEPA representation-alignment loss.
The memory update is driven by the current
observation rather than a future frame.

\subsection{Optimization and Validation}
\label{app:real_robot_optimization}

All three real-robot policies are trained for 2,000 optimizer updates under the same optimization configuration. We use AdamW with global gradient clipping at $1.0$, training on four NVIDIA H100 GPUs
distributed across two nodes. The microbatch size is four per GPU with four gradient-accumulation steps, giving an effective global batch size of 64. Exponential moving averaging (EMA) is disabled.

We monitor held-out action flow-matching validation loss on the fixed validation episodes. The action loss is used for validation monitoring rather than the warm-up-weighted joint objective, whose effective scale changes during the first 500 updates. For a consistent training budget, all three methods are evaluated using the checkpoints obtained after 2,000 optimizer updates, rather than selecting checkpoints based on the lowest validation loss. All methods follow the same real-world evaluation protocol, with model-specific optimization and supervision as described above.

\begin{table}[t]
\centering
\caption{
Real-robot fine-tuning configuration. All three policies share the demonstration dataset, data split, optimizer settings, and 2,000-update training duration; JEPA supervision and TTT memory are enabled only in the corresponding variants.
}
\label{tab:real_robot_training_config}
\footnotesize
\setlength{\tabcolsep}{3pt}
\renewcommand{\arraystretch}{1.07}

\begin{tabular}{
@{}p{0.34\columnwidth}p{0.60\columnwidth}@{}
}
\toprule
\textbf{Component} & \textbf{Configuration} \\
\midrule
Demonstrations
& LeRobot v2.1; 150 episodes; 19,246 frames
at 30\,Hz \\
Train / validation
& 142 / 8 episodes; episode-disjoint split \\
Visual input
& Front and wrist RGB, $640\times480$;
missing right-wrist view masked \\
State / action
& 14-D state; 7-D robot action
(6 joint deltas + absolute gripper) \\
Initialization
& Official $\pi_{0.5}$ base policy \\
Action horizon
& 16 actions; execute first 8 before replanning \\
JEPA teacher
& Frozen V-JEPA 2.1 ViT-g/384 \\

JEPA target
& Jointly encoded current/future front RGB;
offset 31 frames; $576\times1,408$ \\

Auxiliary loss
& Patch-wise cosine alignment;
weight $0\rightarrow0.1$ over 500 updates \\

Memory
& WE; DiT blocks 4, 8, 12, 16 (one-based) \\
Temporal context
& 16 observations; stride 8 frames;
TBPTT length 8 \\
Optimization
& AdamW; global gradient clipping 1.0;
EMA disabled \\
Batch / hardware
& Global batch 64;
4 H100 GPUs across 2 nodes \\
Training duration
& 2,000 optimizer updates per method \\
Offline validation
& Held-out action flow-matching loss \\
\bottomrule
\end{tabular}
\end{table}

\subsection{Real-World Deployment and Evaluation}
\label{app:real_robot_deployment}

For deployment, a single NVIDIA RTX A6000 GPU runs policy inference and, for SReWAM, online TTT memory updates. All three policies use one-step action generation. At each planning cycle, the policy predicts 16 actions, executes the first eight, then receives updated observations and replans. The robot uses the same two RGB camera views and task instruction across methods. A trial is considered successful when the can is placed on the plate.

We evaluate the original tabletop setting used for data collection and a white-tabletop setting that changes the background appearance while preserving the task objective. \emph{The white tabletop does not appear in the training demonstrations of any of the three methods}, making it a held-out visual out-of-distribution (OOD) setting for this tabletop appearance change. No method is fine-tuned on the white-tabletop evaluation trials.

Each policy is evaluated for 30 trials in the original setting and 10 trials in the white-tabletop setting. The $\pi_{0.5}$ baseline succeeds in 25/30 and 4/10 trials, $\pi_{0.5}$ + JEPA Obj. in 28/30 and 8/10, and $\pi_{0.5}$ + SReWAM (WE) in 30/30 and 10/10, respectively, as summarized in Table~\ref{tab:real_robot}. Relative to the JEPA-adapted baseline, SReWAM gains 6.7 and 20.0 percentage points in the original and white-tabletop conditions. These results provide task-specific evidence of improved manipulation performance under the tested appearance shift; the 10-trial OOD evaluation is not sufficient to establish broad robustness across unseen environments.

\end{document}